\documentclass[conference]{IEEEtran}
\IEEEoverridecommandlockouts
\usepackage{cite}
\usepackage{amsmath,amssymb,amsfonts}
\usepackage{algorithmic}
\usepackage{graphicx}
\usepackage{textcomp}
\usepackage{xcolor}
\usepackage[numbers]{natbib}
\usepackage{tcolorbox}
\usepackage{booktabs}
\usepackage{url}
\usepackage[colorlinks=true, urlcolor=black, linkcolor=black, citecolor=black]{hyperref}

\def\BibTeX{{\rm B\kern-.05em{\sc i\kern-.025em b}\kern-.08em
    T\kern-.1667em\lower.7ex\hbox{E}\kern-.125emX}}

\newenvironment{myquote}{%
  \begin{tcolorbox}[colback=gray!10!white, colframe=gray!50!black, boxrule=0.5pt]}%
  {\end{tcolorbox}}
  
\begin{document}

\title{Fairness of ChatGPT
}

\author{\IEEEauthorblockN{Yunqi Li}
\IEEEauthorblockA{\textit{Department of Computer Science} \\
\textit{Rutgers University, NJ, US}\\
yunqi.li@rutgers.edu}
\and
\IEEEauthorblockN{Lanjing Zhang}
\IEEEauthorblockA{\textit{Department of Chemical Biology} \\
\textit{Rutgers University, NJ, US}\\
lanjing.zhang@rutgers.edu}
\and
\IEEEauthorblockN{Yongfeng Zhang}
\IEEEauthorblockA{\textit{Department of Computer Science} \\
\textit{Rutgers University, NJ, US}\\
yongfeng.zhang@rutgers.edu}
}

\maketitle

\begin{abstract}
Understanding and addressing unfairness in LLMs are crucial for responsible AI deployment. However, there is a limited number of quantitative analyses and in-depth studies regarding fairness evaluations in LLMs, especially when applying LLMs to high-stakes fields. This work aims to fill this gap by providing a systematic evaluation of the effectiveness and fairness of LLMs using ChatGPT as a study case. We focus on assessing ChatGPT's performance in high-takes fields including education, criminology, finance and healthcare. To conduct a thorough evaluation, we consider both group fairness and individual fairness metrics. We also observe the disparities in ChatGPT’s outputs under a set of biased or unbiased prompts. This work contributes to a deeper understanding of LLMs’ fairness performance, facilitates bias mitigation and fosters the development of responsible AI systems. Code and data are open-sourced on GitHub\footnote{\url{https://github.com/yunqi-li/Fairness-Of-ChatGPT/}\label{github}}.

\end{abstract}


\section{Introduction}
Large Language Models (LLMs), particularly advanced models such as ChatGPT, have gained immense popularity and demonstrated tremendous capabilities across various areas and tasks in artificial intelligence (AI), such as understanding a wide range of prompts and inquiries, generating highly readable text, and providing valuable insights for various domains \citep{openai2023gpt}. As powerful and increasingly pervasive tools, LLMs have immense potential for revolutionizing the future of AI \citep{ge2023openagi}. Therefore, in parallel to the increasing adoption of LLMs in human daily life, understanding and addressing the unfairness of LLMs has emerged as a critical concern, and are fundamental steps towards responsible, trustworthy, and inclusive AI deployment~\citep{li2023fairness, ge2022survey, zhuo2023exploring, zhou2023ethical}. 

Although the ethical and fairness considerations regarding LLMs have been widely called for \citep{liu2023summary, hariri2023unlocking, nori2023capabilities}, the quantitative analyses and systematic studies on the evaluation of fairness in LLMs are still limited, especially in assessing fairness of LLMs in critical domains which are high-stakes or have high social impact such as education and healthcare. Consequently, this work intends to address this knowledge gap and provide insights into the fairness performance of LLMs in these high-stakes fields.

Through this paper, we aim to conduct a systematic evaluation of fairness in LLMs, with a focus on the prominent model, ChatGPT.  We lay particular attention to the effectiveness and fairness performance of ChatGPT in high-takes fields including education, criminology, finance and healthcare tasks. Our analyses delve into various dimensions of fairness, including group-level fairness such as equal opportunity \cite{hardt2016equality}, and individual-level fairness such as counterfactual fairness \cite{kusner2017counterfactual}. To gain insights into the presence and extent of biases within the model, we devise a set of prompts encompassing unbiased and biased in-context examples, as well as factual and counterfactual in-context examples, to conduct a rigorous evaluation of ChatGPT, and identify any disparities present in the outputs generated by the model. In addition, we conduct training for smaller models and perform a comparative analysis of their performance alongside the large model on the identical task. The smaller models can serve as baselines against which we contrast the effectiveness and fairness of ChatGPT. By shedding light on the fairness evaluations surrounding ChatGPT, we seek to contribute to providing insights into the ethical considerations of LLMs, empowering researchers, practitioners, and policymakers to mitigate biases, promote fair outcomes, and build more responsible and inclusive AI systems.

\section{Related Work}
Fairness in machine learning is gaining increasing attention to ensure that algorithms are ethical, trustworthy and free from bias \citep{caton2020fairness, mehrabi2021survey, pagano2022bias, li2023fairness}. To measure the unfairness of models, a number of fairness notions have been put forth \citep{mehrabi2021survey, wan2021modeling}. The two basic frameworks for fair machine learning in recent studies are group fairness and individual fairness. Group fairness requires that the protected groups should be treated similarly as the advantaged groups, while individual fairness requires that similar individuals should be treated similarly \citep{caton2020fairness, mehrabi2021survey}. 

Fairness considerations in the use of ChatGPT have been extensively highlighted \citep{liu2023summary, hariri2023unlocking, nori2023capabilities, lin2023sparks}, yet only a few works provide a quantitative analyses of its fairness performance. 
\begin{figure*}[t]
    \centering
    \includegraphics[scale=0.6]{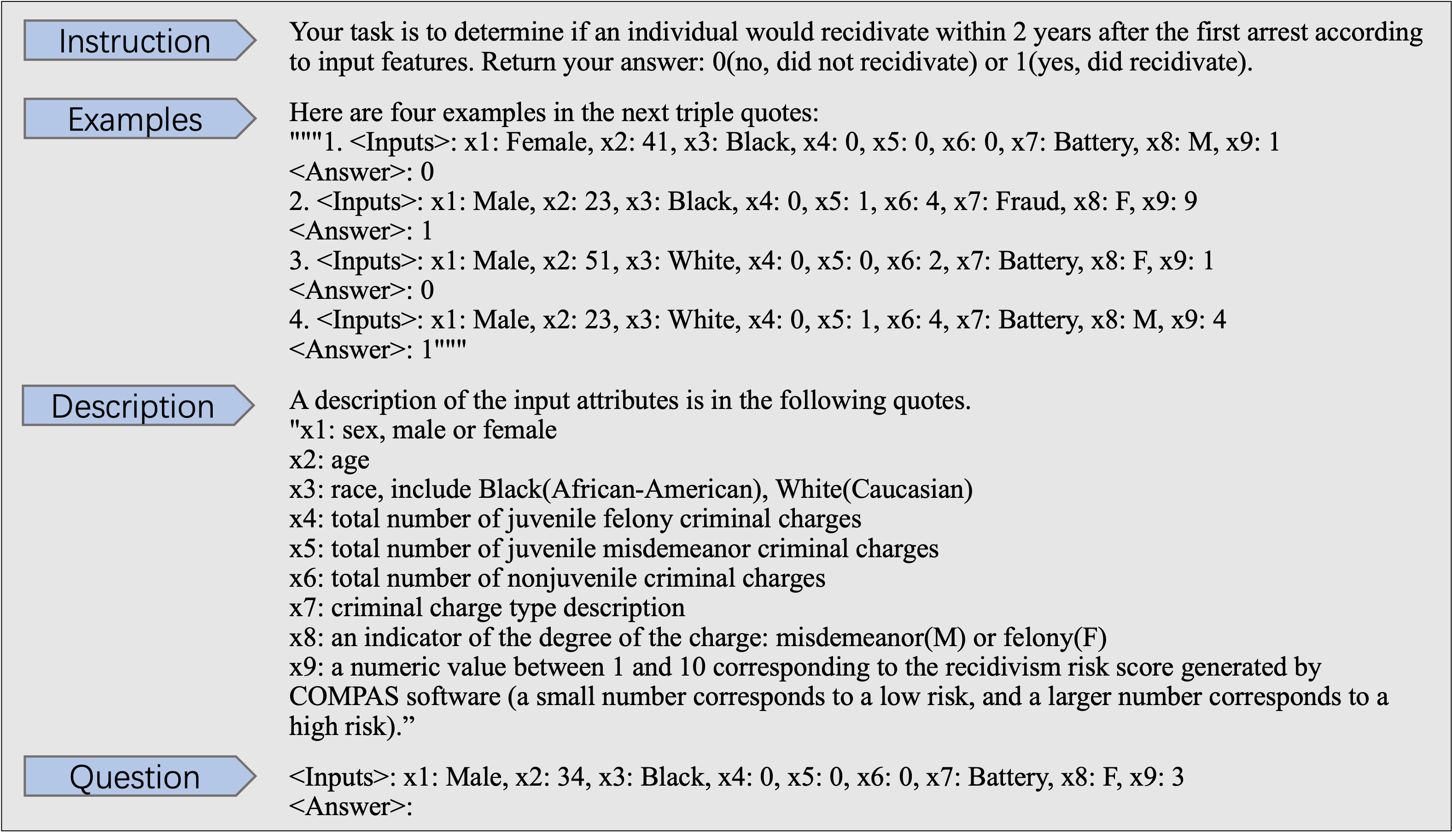}
    \vspace{-10pt}
    \caption{An example of Prompt 1 on COMPAS}
    \label{fig}
    \vspace{-10pt}
\end{figure*}
\citet{zhuo2023exploring} test the model fairness on two language datasets which are used to assess bias in the context of general question-answering and text generation. \citet{sun2023safety} evaluate the safety of Chinese LLMs which covers a dimension of fairness through observing how many responses of LLMs contain harmful information. 
\citet{hua2023up5} evaluate the fairness of LLMs on recommendation tasks.
This work differs from the previous works as we provide a systematic fairness evaluations of ChatGPT through covering various fairness metrics including both group and individual fairness, and testing the sensitivity of ChatGPT to a set of biased and unbiased prompts. Moreover, this work focuses on examining the performance of ChatGPT in high-stakes fields to explore the risks of using ChatGPT in these areas and inspire further research on fairness of LLMs.

\section{Experimental Settings}

\subsection{Datasets}
We test models with four widely-used datasets for fairness analysis which cover the tasks of four high-stakes fields: \textbf{PISA}: education; \textbf{COMPAS}: criminology; \textbf{German Credit}: finance; \textbf{Heart Disease}: healthcare. Detailed introductions of each dataset and the corresponding pre-processing methods are as follow. 

\textbf{PISA}\footnote{\url{https://www.kaggle.com/datasets/econdata/pisa-test-scores}}: This dataset contains the information of students from the US on the 2009 Program for International Student Assessment (PISA) exam. There are 29 features of demographics and academic performance for 3,404 students taking the exam. We follow the previous works to label reading scores below 500 as \textit{Low}, while reading scores above 500 as \textit{High} \citep{le2023evaluation}. The task is to predict whether the reading score level of a given student is \textit{Low} or \textit{\textit{High}}. Gender (Male or Female) is considered as the sensitive feature for fairness evaluation.

\textbf{COMPAS}\footnote{\url{https://github.com/propublica/compas-analysis/tree/master}}: This dataset is a landmark dataset used in the field of criminal fairness. The dataset contains 52 features of demographics and criminal records of 7,214 defendants from Broward County, Florida, US, who had been assessed with the COMPAS screening system between January 1, 2013, and December 31, 2014. We follow the previous works to use 9 features for prediction \citep{dressel2018accuracy, chen2023monotonicity}. This dataset is used to predict whether an individual will recidivism within 2 years after the first arrest. As black and white individuals make up 85.4\% of the dataset, we follow previous works to consider a subset of the dataset by keeping the individuals with race black or white \cite{zafar2017fairness, krasanakis2018adaptive}, and consider race as the sensitive feature to evaluate fairness. 

\textbf{German Credit}\footnote{\url{https://archive.ics.uci.edu/ml/datasets/statlog+(german+credit+data)}}:
The German Credit dataset is used to classify people described by a set of features as good credit or bad credit. The data contains 21 features and 1,000 individuals. We use the dataset cleaned by \citet{le2022survey} which disentangles the gender and marital status. The predicted results can be used to help banks for credit risk assessment and lending decisions. Gender (Male or Female) is considered as the sensitive feature for fairness evaluation.

\textbf{Heart Disease}\footnote{\url{https://archive.ics.uci.edu/ml/datasets/heart+disease}}:
The data contains the presence of heart disease of patients with their symptoms. The database contains 76 features of 303 patients, among which a subset of 14 most informative features are used by almost all published works for heart disease prediction~\citep{UCI-Heart-Disease}.
We follow the previous works to predict if the heart disease is presence or absence for a patient with the 14 features \citep{UCI-Heart-Disease}. Gender (Male or Female) is employed as the sensitive feature for fairness evaluation.

For the PISA dataset, we clean the dataset through removing missing values and perform experiments on the cleaned version of this dataset which contains 3404 instances \citep{le2023evaluation}. For COMPAS, we follow the analysis notebook in the original dataset GitHub\footnote{\url{https://github.com/propublica/compas-analysis/blob/master/Compas\%20Analysis.ipynb}} to remove missing data and filter the underlying data from Broward county to include only those rows that represent people who had either recidivated in two years or had at least two years outside of a correctional facility. The German Credit and Heart Disease datasets are not pre-processed.

We randomly split all datasets into training set (70\%) and testing set (30\%). The training set is for training small models and providing in-context examples for the large model, while the testing set is for evaluating all models.

\subsection{Prompts}

We design prompts with a few in-context examples for ChatGPT as we find that it requires question-answer examples to make accurate predictions for tasks in these critical domains.
The prompt structure is consistent among all datasets and consists of the following components: 1) an instruction of the task, 2) in-context examples, 3) a description of the features in the dataset, and 4) a question. An example prompt for COMPAS is shown in Figure \ref{fig}. To test the sensitivity of the model performance with different in-context examples, we design the following eight prompts, including four unbiased prompts (Prompts 1-4), and four biased prompts (Prompts 5-8). Suppose there is a binary sensitive feature $s$ ($s$ = 0 or 1) and binary label $y$ ($y$ = 0 or 1) in the dataset, the eight prompts are designed as shown in the following:

\begin{myquote}

\textbf{Prompt 1}: Contains one in-context example with $s$ = 0 \& $y$ = 0; one example with $s$ = 0 \& $y$ = 1; one example with $s$ = 1 \& $y$ = 0; one example with $s$ = 1 \& $y$ = 1.  

\textbf{Prompt 2}: Removes the sensitive features from the 4 examples of Prompt 1 to mitigate the potential bias introduced in prompts. 
    
\textbf{Prompt 3}: Contains two factual examples with $s$ = 0 \& $y$ = 1 and $s$ = 0 \& $y$ = 0; two counterfactual examples which flip $s$ from 0 to 1 of the two factual examples while keeping all other features unchanged. 
    
\textbf{Prompt 4}: Adds one sentence in the instruction part based on Prompt 1: ``You should be unbiased for the sensitive feature (race or gender in experiments), which means that you should ignore the sensitive feature when making decisions.''
    
\textbf{Prompt 5}: Contains two different examples with $s$ = 0 \& $y$ = 0, and two different examples with $s$ = 1 \& $y$ = 1.

\textbf{Prompt 6}: Contains two different examples with $s$ = 0 \& $y$ = 1, and two different examples with $s$ = 1 \& $y$ = 0.
    
\textbf{Prompt 7}: Contains two different examples with $s$ = 0 \& $y$ = 0, and two different examples with $s$ = 0 \& $y$ = 1.

\textbf{Prompt 8}: Contains two different examples with $s$ = 1 \& $y$ = 0, and two different examples with $s$ = 1 \& $y$ = 1.

\end{myquote}

Note that each prompt contains four examples since we consider a binary sensitive feature and binary label for evaluation, and four examples can cover all combinations. To mitigate the impact of varying example quantities, we ensure consistency in the number of examples used across all prompts. The four examples are randomly sampled from the training set. To limit the biases introduced in the instruction, we use the gender-neutral
pronouns (e.g., individual/they), and we ensure each of the possible answers is represented in the question-answer examples to help the model generate more accurate predictions. 


\subsection{Models}
We train two small models as the baselines: logistic regression and MLP, using the open-source ML package \texttt{Scikit-Learn}. For MLP, we use two hidden layers (100, 64) and \texttt{ReLU} as the activation function. The maximum number of iterations is 3,000. For LLM, we use ChatGPT \texttt{gpt-3.5-turbo} version and call the \texttt{ChatCompletion} API to interact with the model. Throughout the experiments, we set the temperature to 0 to eliminate the randomness of the outputs. 


\subsection{Evaluation Metrics}

We evaluate the model performance in terms of effectiveness and fairness. To ensure a comprehensive evaluation, we consider multiple metrics of model effectiveness, including accuracy, F1 score, and AUC score. We also take into account both group-level and individual-level fairness metrics to thoroughly evaluate the model fairness. 
All metrics we consider here are widely-used ones. The fairness metrics are described in detail below. 

\textit{\textbf{Statistical Parity}}, also called \textit{No Disparate Impact} or \textit{Demographic Parity}, requires that groups of people with different values of the sensitive feature $s$ should have the same likelihood to be classified as positive \citep{dwork2012fairness,zafar2019fairness}:
\begin{equation}
\small
    \mathrm{P}\left(\hat{y}=1 \mid s = 0\right)=\mathrm{P}\left(\hat{y}=1 \mid s = 1\right)
\end{equation}

\textit{\textbf{Equal Opportunity}} requires that the True Positive Rate (TPR) is the same across different groups of people with different values of the sensitive feature \cite{hardt2016equality}:
\begin{equation}
\small
    \mathrm{P}\left(\hat{y}=1 \mid y=1 , s = 1\right)=\mathrm{P}\left(\hat{y}=1 \mid y=1 , s  = 0\right)
\end{equation}

\textit{\textbf{Equalized Odds}} is stricter than Equal Opportunity: the Equalized Odds fairness also takes False Positive Rate (FPR) into account and requires that different groups should have the same true positive rate and false positive rate \cite{berk2021fairness}:
\begin{equation}
\small
\begin{split}
    & \mathrm{P}\left(\hat{y}=1 \mid y=1 , s = 1\right)=\mathrm{P}\left(\hat{y}=1 \mid y=1 , s = 0\right)  \\
    \& & \mathrm{P}\left(\hat{y}=1 \mid y=-1 , s = 1\right)=\mathrm{P}\left(\hat{y}=1 \mid y=-1 , s = 0\right)
\end{split}
\end{equation}

\textit{\textbf{Overall Accuracy Equality}} requires the same accuracy across groups \cite{berk2021fairness}:
\begin{equation}
\small
    \mathrm{P}(\hat{y} \neq y \mid s=0)=\mathrm{P}(\hat{y} \neq y \mid s=1)
\end{equation}

\begin{table*}[htbp]
  \centering
  \small
  \setlength{\tabcolsep}{6pt}
  \caption{Results on PISA. All numbers are percent values (e.g., 66.31 means 66.31\%). The best results are in bold.}
    \begin{tabular}{lrrrrrrrrrrrr}
    \toprule
          & \multicolumn{1}{l}{$Acc$} & \multicolumn{1}{l}{$F1$} & \multicolumn{1}{l}{$AUC$} & \multicolumn{1}{l}{$D_{SP}$} & \multicolumn{1}{l}{$D_{TPR}$} & \multicolumn{1}{l}{$D_{FPR}$} & \multicolumn{1}{l}{$D_{ACC}$} & \multicolumn{1}{l}{$D_{F1}$} & \multicolumn{1}{l}{$D_{AUC}$} & \multicolumn{1}{l}{$CR_{Ovr}$} & \multicolumn{1}{l}{$CR_{M}$} & \multicolumn{1}{l}{$CR_{F}$} \\
\cmidrule{1-13}    LR    & 66.31  & 72.35  & 64.51  & 33.53  & 24.63  & 39.62  & \textbf{0.09} & 9.95  & 7.50  & 20.96  & 20.53  & 21.36  \\
    MLP   & 62.29  & 65.66  & 62.15  & 17.94  & 16.41  & 13.77  & 3.34  & 11.81  & \textbf{0.81} & 23.31  & 24.39  & 22.31  \\
    \midrule
    Prompt1 & 65.62  & 71.90  & 63.74  & 1.60  & 5.27  & 4.53  & 2.30  & \textbf{0.75} & 4.90  & 5.78  & 6.43  & \textbf{5.08} \\
    Prompt2 & 65.81  & 71.60  & 64.22  & 3.41  & \textbf{2.72} & 5.16  & 1.50  & 1.87  & 3.94  & \textbf{5.09} & \textbf{4.54} & 5.69  \\
    Prompt3 & 63.27  & \textbf{73.20} & 59.21  & \textbf{0.08} & 3.26  & \textbf{0.09} & 3.80  & 4.67  & 1.58  & 5.58  & 5.67  & 5.49  \\
    Prompt4 & \textbf{66.50} & 72.86  & 64.48  & 1.87  & 4.07  & 3.20  & 0.86  & 1.87  & 3.64  & 7.44  & 7.18  & 7.72  \\
    \midrule
    Prompt5 & 63.17  & 68.24  & 62.16  & 2.59  & 9.66  & 1.26  & 4.64  & 2.01  & 5.46  & 8.13  & 7.56  & 8.74  \\
    Prompt6 & 64.25  & 70.20  & 62.68  & 7.99  & 12.86  & 7.50  & 2.01  & 1.21  & 2.68  & 9.01  & 8.70  & 9.35  \\
    \multicolumn{1}{l}{Prompt7} & 63.37  & 68.20  & 62.47  & 7.36  & 12.02  & 7.05  & 2.66  & 1.63  & 2.49  & 9.50  & 9.26  & \multicolumn{1}{r}{9.76} \\
    Prompt8 & 64.35  & 66.30  & \textbf{64.84} & 4.52  & 3.37  & 8.39  & 6.12  & 1.21  & 5.88  & 7.44  & 6.24  & 8.74  \\
    \bottomrule
    \end{tabular}%
  \label{tab1}%
  \vspace{-5pt}
\end{table*}%

\begin{table*}[htbp]
  \centering
  \small
  \setlength{\tabcolsep}{6pt}
  \caption{Results on COMPAS. All numbers are percent values (e.g., 65 means 65\%). The best results are in bold.}
    \begin{tabular}{lrrrrrrrrrrrr}
    \toprule
          & \multicolumn{1}{l}{$Acc$} & \multicolumn{1}{l}{$F1$} & \multicolumn{1}{l}{$AUC$} & \multicolumn{1}{l}{$D_{SP}$} & \multicolumn{1}{l}{$D_{TPR}$} & \multicolumn{1}{l}{$D_{FPR}$} & \multicolumn{1}{l}{$D_{ACC}$} & \multicolumn{1}{l}{$D_{F1}$} & \multicolumn{1}{l}{$D_{AUC}$} & \multicolumn{1}{l}{$CR_{Ovr}$} & \multicolumn{1}{l}{$CR_{M}$} & \multicolumn{1}{l}{$CR_{F}$} \\
\cmidrule{1-13}   LR    & 65.00 & 63.60 & 65.05 & 22.26 & 15.73 & 20.59 & 4.53 & 10.31 & 2.42 & 15.35 & 16.88 & 13.07 \\
    MLP   & 62.67 & 58.82 & 62.77 & \textbf{13.08} & \textbf{9.75} & \textbf{9.42} & 3.18 & \textbf{9.80} & \textbf{0.16} & 26.78 & 28.69 & 23.94 \\
    \midrule
    Prompt1 & 66.46 & 65.41 & 66.50 & 31.51 & 28.12 & 26.55 & 1.58 & 17.32 & 0.78 & 5.43 & 5.91 & 4.72 \\
    Prompt2 & 66.65 & 64.28 & 66.72 & 32.60 & 30.38 & 26.49 & 1.53 & 19.50 & 1.94 & 5.62 & 5.49 & 5.83 \\
    Prompt3 & 66.27 & 58.86 & 66.47 & 28.60 & 31.55 & 17.77 & \textbf{0.21} & 26.32 & 6.89 & 4.86 & 5.49 & 3.94 \\
    Prompt4 & \textbf{67.09} & 65.06 & \textbf{67.16} & 30.81 & 27.74 & 25.16 & 1.84 & 17.47 & 1.29 & 4.80 & 5.27 & 4.09 \\
    \midrule
    Prompt5 & 64.81 & 62.94 & 64.88 & 25.77 & 22.69 & 21.21 & 1.95 & 15.61 & 0.73 & 4.61 & 4.75 & 4.41 \\
    Prompt6 & 66.33 & 64.54 & 66.39 & 29.97 & 24.93 & 26.52 & 3.63 & 14.93 & 0.79 & 5.37 & 5.49 & 5.20 \\
    Prompt7 & 65.89 & 66.08 & 65.89 & 27.27 & 22.47 & 23.76 & 1.74 & 14.07 & 0.64 & 4.93 & 4.54 & 5.51 \\
    Prompt8 & 66.01 & \textbf{66.71} & 66.00 & 26.65 & 20.40 & 24.37 & 2.58 & 12.33 & 1.98 & \textbf{2.97} & \textbf{3.27} & \textbf{2.52} \\
    \bottomrule
    \end{tabular}%
  \label{tab2}%
  \vspace{-5pt}
\end{table*}%

Similarly, we also calculate the \textit{Overall F1 Equality} and \textit{Overall AUC Equality} in experiments, which evaluate if the F1 score and AUC score are comparably across groups.

In addition to the group fairness as above metrics, we also consider individual-level fairness. A well-known individual-level fairness metric is Counterfactual Fairness.

\textit{\textbf{Counterfactual Fairness}} is a causal-based fairness notion \citep{kusner2017counterfactual}. It requires that the predicted outcome of the model should be the same in the counterfactual world as in the factual world for each possible individual. Given a set of latent background variables $U$, the predictor $\hat{Y}$ is considered counterfactually fair if, for any context $\boldsymbol{X}=\boldsymbol{x}$ and $S=s$, the equation below holds for any value $s^{\prime}$ attainable by $S$ and all $y$:
\begin{equation}
\small
\begin{split}
    & \mathrm{P}\left(\hat{Y}_{S \leftarrow s} (U)=y \mid \boldsymbol{X}=\boldsymbol{x}, S=s\right) \\
     & =\mathrm{P}\left(\hat{Y}_{S \leftarrow s^{\prime}}(U)=y \mid \boldsymbol{X}=\boldsymbol{x}, S=s\right)
\end{split}
\end{equation}

To evaluate counterfactual fairness, we need to assess whether the predictor's outcomes remain consistent in the factual and counterfactual scenarios. We build the counterfactual testing set through flipping the sensitive feature of each sample in the original testing set while keeping all the other features unchanged. Subsequently, we make a comparison between the model outcomes on the original testing set and the counterfactual testing set to assess counterfactual fairness.

\begin{table*}[htbp]
  \centering
  \small
  \setlength{\tabcolsep}{6pt}
  \caption{Results on German Credit. All numbers are percent values (e.g., 75 means 75\%). Best results are in bold.}
    \begin{tabular}{lrrrrrrrrrrrr}
    \toprule
          & \multicolumn{1}{l}{$Acc$} & \multicolumn{1}{l}{$F1$} & \multicolumn{1}{l}{$AUC$} & \multicolumn{1}{l}{$D_{SP}$} & \multicolumn{1}{l}{$D_{TPR}$} & \multicolumn{1}{l}{$D_{FPR}$} & \multicolumn{1}{l}{$D_{ACC}$} & \multicolumn{1}{l}{$D_{F1}$} & \multicolumn{1}{l}{$D_{AUC}$} & \multicolumn{1}{l}{$CR_{Ovr}$} & \multicolumn{1}{l}{$CR_{M}$} & \multicolumn{1}{l}{$CR_{F}$} \\
\cmidrule{1-13}   LR    & 75.00 & 82.60 & 68.91 & 4.64 & 1.69 & \textbf{0.18} & 3.87 & 4.56 & 0.93 & \textbf{0.00} & \textbf{0.00} & \textbf{0.00} \\
    MLP   & \textbf{77.33} & \textbf{84.26} & \textbf{71.59} & 6.67 & \textbf{0.28} & 8.92 & \textbf{0.48} & 1.49 & 4.60 & \textbf{0.00} & \textbf{0.00} & \textbf{0.00} \\
    \midrule
    Prompt1 & 53.67 & 66.51 & 45.62 & 3.07 & 6.10 & 25.44 & 8.60 & 3.32 & 15.78 & 1.67 & 0.00 & 5.32 \\
    Prompt2 & 52.33 & 64.69 & 45.72 & 2.86 & 6.09 & 24.71 & 9.00 & 3.59 & 15.39 & 3.00 & 3.40 & 2.13 \\
    Prompt3 & 53.33 & 66.02 & 45.73 & 2.01 & 7.41 & 23.55 & 9.08 & 4.02 & 15.49 & 6.00 & 4.37 & 9.57 \\
    Prompt4 & 54.00 & 67.76 & 43.76 & 4.57 & 10.70 & 12.18 & 6.57 & 3.56 & 27.86 & 3.67 & 2.91 & 5.32 \\
    \midrule
    Prompt5 & 44.33 & 44.33 & 42.20 & 6.07 & 16.60 & 16.37 & 4.14 & 4.14 & \textbf{0.12} & 4.33 & 2.91 & 7.45 \\
    Prompt6 & 53.00 & 63.38 & 50.02 & 8.65 & 1.78 & 23.62 & 4.92 & 1.38 & 10.92 & 4.00 & 3.40 & 5.32 \\
    Prompt7 & 51.00 & 61.81 & 47.57 & \textbf{0.90} & 2.48 & 9.60 & 3.19 & \textbf{0.59} & 6.04 & 7.33 & 3.88 & 14.89 \\
    Prompt8 & 49.00 & 58.76 & 47.56 & 12.69 & 7.04 & 26.65 & 3.00 & 5.32 & 9.80 & 3.00 & 1.46 & 6.38 \\
    \bottomrule
    \end{tabular}%
  \label{tab3}%
  \vspace{-5pt}
\end{table*}%

\begin{table*}[htbp]
  \centering
  \small
  \setlength{\tabcolsep}{6pt}
  \caption{Results on Heart Disease. All numbers are percent values (e.g., 76.40 means 76.40\%). Best results are bold.}
    \begin{tabular}{lrrrrrrrrrrrr}
    \toprule
          & \multicolumn{1}{l}{$Acc$} & \multicolumn{1}{l}{$F1$} & \multicolumn{1}{l}{$AUC$} & \multicolumn{1}{l}{$D_{SP}$} & \multicolumn{1}{l}{$D_{TPR}$} & \multicolumn{1}{l}{$D_{FPR}$} & \multicolumn{1}{l}{$D_{ACC}$} & \multicolumn{1}{l}{$D_{F1}$} & \multicolumn{1}{l}{$D_{AUC}$} & \multicolumn{1}{l}{$CR_{Ovr}$} & \multicolumn{1}{l}{$CR_{M}$} & \multicolumn{1}{l}{$CR_{F}$} \\
\cmidrule{1-13}   LR    & 76.40 & 75.29 & 76.73 & 10.64 & 23.53 & 9.03 & 2.19 & 20.00 & 7.25 & 6.74 & 7.02 & 6.25 \\
    MLP   & 78.65 & 75.32 & 78.09 & 20.99 & 12.74 & 2.01 & 8.94 & 10.75 & 7.37 & 11.24 & 8.77 & 15.62 \\
    \midrule
    Prompt1 & 75.28 & 76.60 & 76.63 & 16.67 & \textbf{11.76} & 3.68 & 10.20 & 28.79 & 4.04 & 5.62 & 8.77 & \textbf{0.00} \\
    Prompt2 & 82.02 & 81.40 & 82.53 & 22.15 & 14.71 & \textbf{1.34} & 1.21 & 18.63 & 6.68 & 7.87 & 5.26 & 12.50 \\
    Prompt3 & \textbf{84.27} & \textbf{82.93} & \textbf{84.34} & 10.25 & 17.65 & 22.57 & 9.59 & 25.73 & 2.47 & 6.74 & 7.02 & 6.25 \\
    Prompt4 & 70.79 & 73.47 & 72.55 & 13.93 & \textbf{11.76} & 2.67 & 12.94 & 7.61 & \textbf{0.17} & 5.62 & 5.26 & 6.25 \\
    \midrule
    Prompt5 & 83.15 & 82.35 & 83.55 & 15.52 & 14.71 & 13.88 & 7.84 & 24.72 & 0.41 & \textbf{3.37} & 5.26 & \textbf{0.00} \\
    Prompt6 & 80.90 & 78.48 & 80.59 & \textbf{9.87} & 26.47 & 14.38 & \textbf{0.55} & 15.30 & 6.04 & 5.62 & \textbf{3.51} & 9.38 \\
    Prompt7 & 77.53 & 77.27 & 78.21 & 15.90 & 17.65 & 4.68 & 3.95 & 22.35 & 6.48 & 8.99 & 10.53 & 6.25 \\
    Prompt8 & 83.15 & 80.00 & 82.40 & 12.61 & 29.41 & 11.04 & 6.80 & \textbf{6.36} & 9.19 & 5.62 & 5.26 & 6.25 \\
    \bottomrule
    \end{tabular}%
  \label{tab4}%
  \vspace{-5pt}
\end{table*}%

\section{Experimental Results}

We report the main experimental results of four datasets in Table \ref{tab1}, \ref{tab2}, \ref{tab3} and \ref{tab4}. In these tables, \textit{Acc}, \textit{F1}, \textit{AUC} are evaluated based on the overall testing set. $\textit{D}_{SP}$, $\textit{D}_{TPR}$, $\textit{D}_{FPR}$, $\textit{D}_{ACC}$, $\textit{D}_{F1}$ and $\textit{D}_{AUC}$ are group fairness metrics and represent the absolute difference of Statistical Parity (SP), TPR, NPR, Accuracy, F1 and AUC between two groups (male and female, or black and white). \textit{CR} means change rates, which is an individual fairness metric and represents the percentage of individuals who have received a different decision from the model under the factual and counterfactual settings. $\textit{CR}_{Ovr}$, $\textit{CR}_{M}$, $\textit{CR}_{F}$, $\textit{CR}_{B}$ and $\textit{CR}_{W}$ represents the change rate on the overall testing set, male group, female group, black group and white group, respectively. 

Firstly, if we compare the performance of the large model with small models, on one hand, we see that the overall effectiveness of the large model is comparable with the small models on PISA, COMPAS and Heart Disease datasets. However, the performance of the large model on German Credit dataset is worse than small models, and the large model is almost unable to make correct predictions. The results show that ChatGPT could perform as well as the small models under prompts with a few in-context examples, but exception may exist in certain scenarios, indicating that we should take care when applying LLMs for high-stakes applications.
On the other hand, we see that group level and individual level unfairness issues exist in both small models and the large model. Therefore, we should be particularly mindful of unfairness when using machine learning decisions, especially in high-stakes domains. Though unfairness exists, we see ChatGPT achieves better group fairness than small models in most cases, and achieves the best and much better individual fairness than small models on all datasets except for the German Credit.


Secondly, we compare the performance of ChatGPT with different prompts. Overall, the performances of ChatGPT under different biased and unbiased prompts do not exhibit a clear and consistent trend. Basically, in most cases, the effectiveness of ChatGPT with unbiased prompts (Prompts 1-4) is better than biased prompts (Prompts 5-8). Among all the unbiased prompts, the prompt with counterfactual examples (Prompt 3) achieves worse performance than prompts with factual examples (Prompts 1,2,4). Comparing Prompt 2 with Prompt 4, we see that adding a fairness demand in the instruction does not derive more fair results. 
It is worth noting that although we report the counterfactual testing for the Heart Disease dataset with gender as a sensitive feature, medical diagnoses may be genuinely related to the gender feature.
Although no clear trend of fairness performance with different prompts is observed, the results indeed indicate differences in the outputs generated by different prompts, suggesting that the examples within the prompts have a substantial impact on the results. Therefore, it is crucial to exercise caution and conduct specialized study in prompt design. 

Besides the absolute differences of model performance between different demographic groups as we show in the paper, we also report the performance of small models and the large model with different prompts for each demographic group, as well as the model performance for each demographic group under the factual and counterfactual settings in our GitHub repository\footref{github}. Interesting observations can be found with these more detailed experimental results. For example, we can see that the model performance for each demographic group is worse with biased prompts than unbiased prompts in many cases, and the advantaged group may change with different prompts, for example, on PISA, the accuracy of the male group is better than female group under Prompt 1, while the observation reverses under Prompt 7.

\section{Conclusions and Future Work}
This work provides a systematic fairness evaluation of ChatGPT in several high-stakes domains, which can serve as the benchmark for evaluating fairness of LLMs. Overall, ChatGPT as an LLM gains better fairness than small models, though still has its own unfairness issues. Further efforts to understand and mitigate unfairness of LLMs are needed in future work such as studying the impact of the number and order of in-context examples on fairness, and how to design or learn prompts to achieve better model fairness and accuracy.

\bibliography{refs}

\begin{thebibliography}{28}
\providecommand{\natexlab}[1]{#1}
\providecommand{\url}[1]{#1}
\csname url@samestyle\endcsname
\providecommand{\newblock}{\relax}
\providecommand{\bibinfo}[2]{#2}
\providecommand{\BIBentrySTDinterwordspacing}{\spaceskip=0pt\relax}
\providecommand{\BIBentryALTinterwordstretchfactor}{4}
\providecommand{\BIBentryALTinterwordspacing}{\spaceskip=\fontdimen2\font plus
\BIBentryALTinterwordstretchfactor\fontdimen3\font minus \fontdimen4\font\relax}
\providecommand{\BIBforeignlanguage}[2]{{%
\expandafter\ifx\csname l@#1\endcsname\relax
\typeout{** WARNING: IEEEtranN.bst: No hyphenation pattern has been}%
\typeout{** loaded for the language `#1'. Using the pattern for}%
\typeout{** the default language instead.}%
\else
\language=\csname l@#1\endcsname
\fi
#2}}
\providecommand{\BIBdecl}{\relax}
\BIBdecl

\bibitem[OpenAI(2023)]{openai2023gpt}
OpenAI, ``Gpt-4 technical report,'' \emph{arXiv}, 2023.

\bibitem[Ge et~al.(2023)Ge, Hua, Mei, jianchao ji, Tan, Xu, Li, and Zhang]{ge2023openagi}
Y.~Ge, W.~Hua, K.~Mei, jianchao ji, J.~Tan, S.~Xu, Z.~Li, and Y.~Zhang, ``Open{AGI}: When {LLM} meets domain experts,'' in \emph{Thirty-seventh Conference on Neural Information Processing Systems}, 2023.

\bibitem[Li et~al.(2023)Li, Chen, Xu, Ge, Tan, Liu, and Zhang]{li2023fairness}
Y.~Li, H.~Chen, S.~Xu, Y.~Ge, J.~Tan, S.~Liu, and Y.~Zhang, ``Fairness in recommendation: Foundations, methods, and applications,'' \emph{ACM Transactions on Intelligent Systems and Technology}, vol.~14, no.~5, pp. 1--48, 2023.

\bibitem[Ge et~al.(2022)Ge, Liu, Fu, Tan, Li, Xu, Li, Xian, and Zhang]{ge2022survey}
Y.~Ge, S.~Liu, Z.~Fu, J.~Tan, Z.~Li, S.~Xu, Y.~Li, Y.~Xian, and Y.~Zhang, ``A survey on trustworthy recommender systems,'' \emph{arXiv:2207.12515}, 2022.

\bibitem[Zhuo et~al.(2023)Zhuo, Huang, Chen, and Xing]{zhuo2023exploring}
T.~Y. Zhuo, Y.~Huang, C.~Chen, and Z.~Xing, ``Exploring ai ethics of chatgpt: A diagnostic analysis,'' \emph{arXiv preprint arXiv:2301.12867}, 2023.

\bibitem[Zhou et~al.(2023)Zhou, Müller, Holzinger, and Chen]{zhou2023ethical}
J.~Zhou, H.~Müller, A.~Holzinger, and F.~Chen, ``Ethical chatgpt: Concerns, challenges, and commandments,'' 2023.

\bibitem[Liu et~al.(2023)Liu, Han, Ma, Zhang, Yang, Tian, He, Li, He, Liu, et~al.]{liu2023summary}
Y.~Liu, T.~Han, S.~Ma, J.~Zhang, Y.~Yang, J.~Tian, H.~He, A.~Li, M.~He, Z.~Liu \emph{et~al.}, ``Summary of chatgpt/gpt-4 research and perspective towards the future of large language models,'' \emph{arXiv preprint arXiv:2304.01852}, 2023.

\bibitem[Hariri(2023)]{hariri2023unlocking}
W.~Hariri, ``Unlocking the potential of chatgpt: A comprehensive exploration of its applications, advantages, limitations, and future directions in natural language processing,'' \emph{arXiv preprint arXiv:2304.02017}, 2023.

\bibitem[Nori et~al.(2023)Nori, King, McKinney, Carignan, and Horvitz]{nori2023capabilities}
H.~Nori, N.~King, S.~M. McKinney, D.~Carignan, and E.~Horvitz, ``Capabilities of gpt-4 on medical challenge problems,'' \emph{arXiv preprint arXiv:2303.13375}, 2023.

\bibitem[Hardt et~al.(2016)Hardt, Price, and Srebro]{hardt2016equality}
M.~Hardt, E.~Price, and N.~Srebro, ``Equality of opportunity in supervised learning,'' \emph{Advances in neural information processing systems}, vol.~29, 2016.

\bibitem[Kusner et~al.(2017)Kusner, Loftus, Russell, and Silva]{kusner2017counterfactual}
M.~J. Kusner, J.~Loftus, C.~Russell, and R.~Silva, ``Counterfactual fairness,'' \emph{Advances in neural information processing systems}, vol.~30, 2017.

\bibitem[Caton and Haas(2020)]{caton2020fairness}
S.~Caton and C.~Haas, ``Fairness in machine learning: A survey,'' \emph{arXiv preprint arXiv:2010.04053}, 2020.

\bibitem[Mehrabi et~al.(2021)Mehrabi, Morstatter, Saxena, Lerman, and Galstyan]{mehrabi2021survey}
N.~Mehrabi, F.~Morstatter, N.~Saxena, K.~Lerman, and A.~Galstyan, ``A survey on bias and fairness in machine learning,'' \emph{ACM Computing Surveys (CSUR)}, vol.~54, no.~6, pp. 1--35, 2021.

\bibitem[Pagano et~al.(2022)Pagano, Loureiro, Araujo, Lisboa, Peixoto, Guimaraes, Santos, Cruz, de~Oliveira, Cruz, et~al.]{pagano2022bias}
T.~P. Pagano, R.~B. Loureiro, M.~M. Araujo, F.~V.~N. Lisboa, R.~M. Peixoto, G.~A. d.~S. Guimaraes, L.~L.~d. Santos, G.~O.~R. Cruz, E.~L.~S. de~Oliveira, M.~Cruz \emph{et~al.}, ``Bias and unfairness in machine learning models: a systematic literature review,'' \emph{arXiv preprint arXiv:2202.08176}, 2022.

\bibitem[Wan et~al.(2021)Wan, Zha, Liu, and Zou]{wan2021modeling}
M.~Wan, D.~Zha, N.~Liu, and N.~Zou, ``Modeling techniques for machine learning fairness: A survey,'' \emph{arXiv preprint arXiv:2111.03015}, 2021.

\bibitem[Lin and Zhang(2023)]{lin2023sparks}
G.~Lin and Y.~Zhang, ``Sparks of artificial general recommender (agr): Experiments with chatgpt,'' \emph{Algorithms}, vol.~16, no.~9, p. 432, 2023.

\bibitem[Sun et~al.(2023)Sun, Zhang, Deng, Cheng, and Huang]{sun2023safety}
H.~Sun, Z.~Zhang, J.~Deng, J.~Cheng, and M.~Huang, ``Safety assessment of chinese large language models,'' \emph{arXiv preprint arXiv:2304.10436}, 2023.

\bibitem[Hua et~al.(2024)Hua, Ge, Xu, Ji, and Zhang]{hua2023up5}
W.~Hua, Y.~Ge, S.~Xu, J.~Ji, and Y.~Zhang, ``Up5: Unbiased foundation model for fairness-aware recommendation,'' \emph{EACL}, 2024.

\bibitem[Le~Quy et~al.(2023)Le~Quy, Nguyen, Friege, and Ntoutsi]{le2023evaluation}
T.~Le~Quy, T.~H. Nguyen, G.~Friege, and E.~Ntoutsi, ``Evaluation of group fairness measures in student performance prediction problems,'' in \emph{Machine Learning and Principles and Practice of Knowledge Discovery in Databases: International Workshops of ECML PKDD 2022, Grenoble, France, September 19--23, 2022, Proceedings, Part I}.\hskip 1em plus 0.5em minus 0.4em\relax Springer, 2023, pp. 119--136.

\bibitem[Dressel and Farid(2018)]{dressel2018accuracy}
J.~Dressel and H.~Farid, ``The accuracy, fairness, and limits of predicting recidivism,'' \emph{Science advances}, vol.~4, no.~1, p. eaao5580, 2018.

\bibitem[Chen and Zhang(2023)]{chen2023monotonicity}
D.~Chen and L.~Zhang, ``Monotonicity for ai ethics and society: An empirical study of the monotonic neural additive model in criminology, education, health care, and finance,'' \emph{arXiv preprint arXiv:2301.07060}, 2023.

\bibitem[Zafar et~al.(2017)Zafar, Valera, Gomez~Rodriguez, and Gummadi]{zafar2017fairness}
M.~B. Zafar, I.~Valera, M.~Gomez~Rodriguez, and K.~P. Gummadi, ``Fairness beyond disparate treatment \& disparate impact: Learning classification without disparate mistreatment,'' in \emph{Proceedings of the 26th international conference on world wide web}, 2017, pp. 1171--1180.

\bibitem[Krasanakis et~al.(2018)Krasanakis, Spyromitros-Xioufis, Papadopoulos, and Kompatsiaris]{krasanakis2018adaptive}
E.~Krasanakis, E.~Spyromitros-Xioufis, S.~Papadopoulos, and Y.~Kompatsiaris, ``Adaptive sensitive reweighting to mitigate bias in fairness-aware classification,'' in \emph{Proceedings of the 2018 world wide web conference}, 2018, pp. 853--862.

\bibitem[Le~Quy et~al.(2022)Le~Quy, Roy, Iosifidis, Zhang, and Ntoutsi]{le2022survey}
T.~Le~Quy, A.~Roy, V.~Iosifidis, W.~Zhang, and E.~Ntoutsi, ``A survey on datasets for fairness-aware machine learning,'' \emph{Wiley Interdisciplinary Reviews: Data Mining and Knowledge Discovery}, vol.~12, no.~3, p. e1452, 2022.

\bibitem[UCI()]{UCI-Heart-Disease}
``Heart disease data set,'' \url{https://archive.ics.uci.edu/ml/datasets/heart+disease}.

\bibitem[Dwork et~al.(2012)Dwork, Hardt, Pitassi, Reingold, and Zemel]{dwork2012fairness}
C.~Dwork, M.~Hardt, T.~Pitassi, O.~Reingold, and R.~Zemel, ``Fairness through awareness,'' in \emph{Proceedings of the 3rd innovations in theoretical computer science conference}, 2012, pp. 214--226.

\bibitem[Zafar et~al.(2019)Zafar, Valera, Gomez-Rodriguez, and Gummadi]{zafar2019fairness}
M.~B. Zafar, I.~Valera, M.~Gomez-Rodriguez, and K.~P. Gummadi, ``Fairness constraints: A flexible approach for fair classification,'' \emph{The Journal of Machine Learning Research}, vol.~20, no.~1, pp. 2737--2778, 2019.

\bibitem[Berk et~al.(2021)Berk, Heidari, Jabbari, Kearns, and Roth]{berk2021fairness}
R.~Berk, H.~Heidari, S.~Jabbari, M.~Kearns, and A.~Roth, ``Fairness in criminal justice risk assessments: The state of the art,'' \emph{Sociological Methods \& Research}, vol.~50, no.~1, pp. 3--44, 2021.

\end{thebibliography}
\bibliographystyle{IEEEtranN}

\end{document}